\documentclass[letterpaper, 10 pt, journal, twoside]{IEEEtran}
\usepackage{amsmath,amsfonts}
\usepackage{algorithmic}
\usepackage{array}
\usepackage[caption=false,font=normalsize,labelfont=sf,textfont=sf]{subfig}
\usepackage{textcomp}
\usepackage{stfloats}
\usepackage{url}
\usepackage{verbatim}
\usepackage{graphicx}
\def\BibTeX{{\rm B\kern-.05em{\sc i\kern-.025em b}\kern-.08em
    T\kern-.1667em\lower.7ex\hbox{E}\kern-.125emX}}
\usepackage{balance}
\usepackage{booktabs}
\usepackage{threeparttable}
\usepackage{multirow}
\usepackage{makecell}
\usepackage{hyperref}   
\usepackage{pifont}
\usepackage[table]{xcolor}
\usepackage{array}
\usepackage{tabularx}

\begin{document}

\title{ImitDiff: Transferring Foundation-Model Priors for Distraction -Robust Visuomotor Policy}

\author{{Yuhang Dong}$^{1}$, {Haizhou Ge}$^{2}$, {Yupei Zeng}$^{1}$, {Jiangning Zhang}$^{3}$, {Beiwen Tian}$^{2}$, {Hongrui Zhu}$^{1}$, {Yufei Jia}$^{2}$, {Ruixiang Wang}$^{4}$, {Zhucun Xue}$^{5}$, {Guyue Zhou}$^{2}$, {Longhua Ma}$^{1}$, {Guanzhong Tian}$^{1}$
\thanks{Manuscript received: May 1, 2025; Revised August 18, 2025; Accepted September 20, 2025. This paper was recommended for publication by Editor Asfour Tamim and Faust Aleksandra upon evaluation of the Associate Editor and Reviewers' comments. This work is supported in part by the National Natural Science Foundation of China under Grant 62303405, in part by Ningbo Natural Science Foundation Project under Grant 2023J400, and in part by Open Research Fund Program of Beijing National Research Center for Information Science and Technology (Corresponding author: Guanzhong Tian and Jiangning Zhang).}
\thanks{$^{1}$Yuhang Dong, Yupei Zeng, Hongrui Zhu, Longhua Ma and Guanzhong Tian are with Ningbo Global Innovation Center, Zhejiang University, Ningbo 315199, China.
    {\tt\footnotesize 22360407@zju.edu.cn; yupeizeng@zju.edu.cn; 22460535@zju.edu.cn; lhma\_zju@zju.edu.cn; gztian@zju.edu.cn.}}
\thanks{$^{2}$Haizhou Ge, Beiwen Tian, Yufei Jia and Guyue Zhou are with Tsinghua University, Beijing 100084, China.
    {\tt\footnotesize ghz23@mails.tsinghua.edu.cn; tbw18@mails.tsinghua.edu.cn;
    jyf23@mails.tsinghua.edu.cn; zhouguyue@air.tsinghua.edu.cn.}}
\thanks{$^{3}$Jiangning Zhang is with Youtu, Tencent, Shanghai 200233, China.
    {\tt\footnotesize 186368@zju.edu.cn.}}
\thanks{$^{4}$Ruixiang Wang is with The Chinese University of Hong Kong, Shenzhen 518172, China.
    {\tt\footnotesize 225040514@link.cuhk.edu.cn.}}
\thanks{$^{5}$Zhucun Xue is with College of Control Science and Engineering, Zhejiang University, Hangzhou 310027, China.
    {\tt\footnotesize 12432038@zju.edu.cn.}}
\thanks{Project page will be available at https://yuhangdong-zju.github.io/ImitDiff/}
\thanks{Digital Object Identifier (DOI): see top of this page.}
}

\markboth{IEEE Robotics and Automation Letters. Preprint Version. Accepted September, 2025}
{Yuhang Dong \MakeLowercase{\textit{et al.}}: ImitDiff} 

\maketitle

\begin{abstract}

\label{abstract}
Visuomotor imitation learning policies enable robots to efficiently acquire manipulation skills from visual demonstrations. However, as scene complexity and visual distractions increase, policies that perform well in simple settings often experience substantial performance degradation. To address this challenge, we propose ImitDiff, a diffusion-based imitation learning policy guided by fine-grained semantics within a dual-resolution workflow. Leveraging pretrained priors of vision-language foundation models, our method transforms high-level instructions into pixel-level visual semantic masks. These masks guide a dual-resolution perception pipeline that captures both global context (e.g., overall layout) from low-resolution observation and fine-grained local features (e.g., geometric details) from high-resolution observation, enabling the policy to focus on task-relevant regions. Additionally, we introduce a consistency-driven diffusion transformer action head that bridges visual semantic conditions and real-time action generation. Extensive experiments demonstrate that ImitDiff outperforms state-of-the-art vision-language manipulation frameworks, as well as visuomotor imitation learning policies, particularly under increased scene complexity and visual distractions. Notably, ImitDiff exhibits strong generalization in zero-shot settings involving novel objects and visual distractions. Furthermore, our consistency-driven action head achieves an order-of-magnitude improvement in inference speed while maintaining competitive success rates.

\end{abstract}


\begin{figure}[t]
    \centering
    \includegraphics[width=1.0\linewidth]{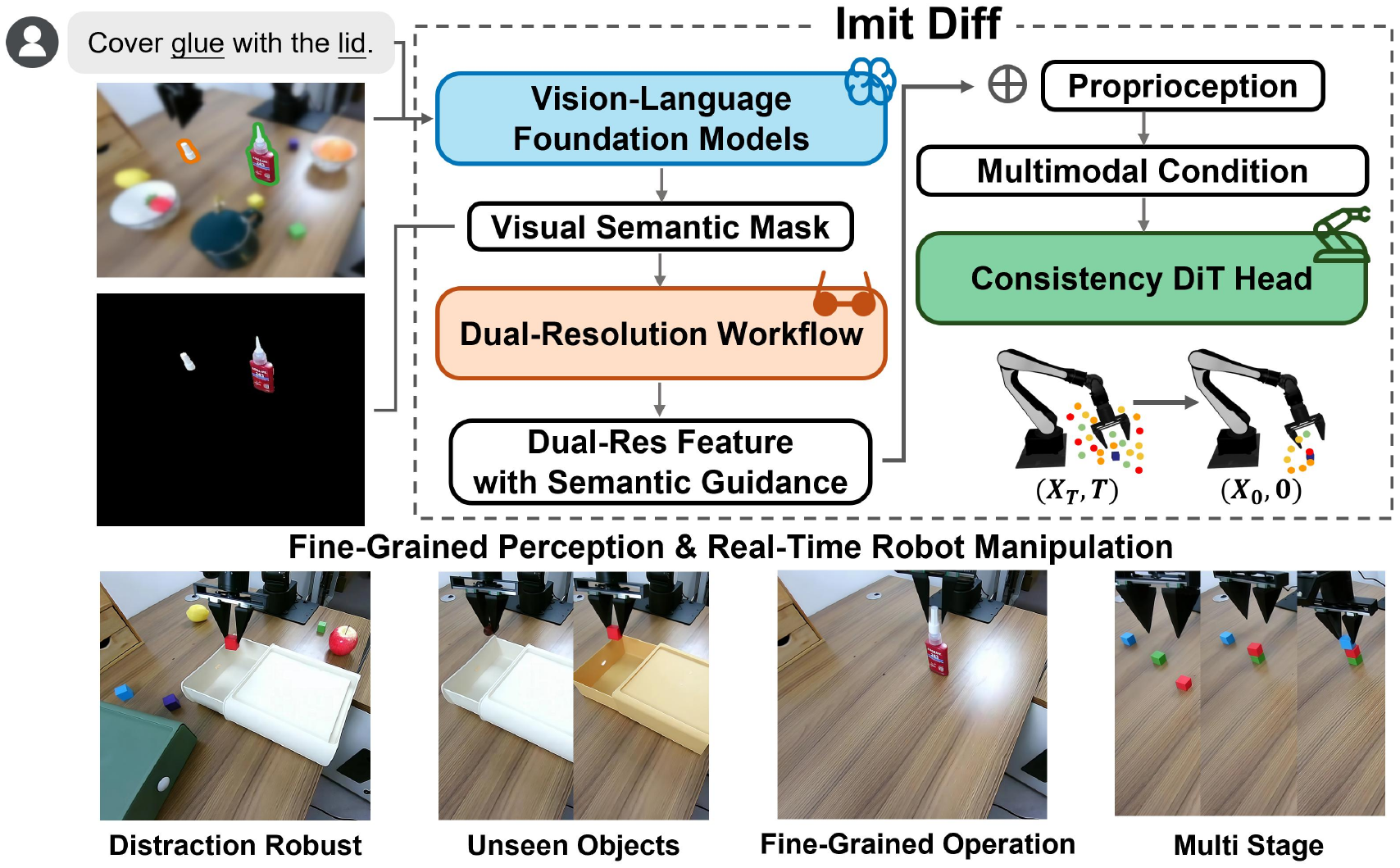} 
    \caption{\textbf{ImitDiff} leverages the priors of vision-language foundation models to transform high-level instructions into pixel-level semantic masks, which finely constrain dual-resolution visual features within the dual-resolution workflow. Based on this, the consistency-driven diffusion transformer (DiT) action head generates executable trajectories in real time under conditional supervision.}
    \label{Fig:Teaser}
\vspace{-1.5\baselineskip}
\end{figure}

\section{Introduction}
\label{Intro}
\IEEEPARstart{A}{chieving} robust and generalizable robotic manipulation in complex environments remains a central and enduring research challenge \cite{UMI}. As a mainstream paradigm for robotic control, visuomotor imitation learning policies enable robots to acquire state estimation and decision-making capabilities from high-dimensional visual and proprioceptive inputs \cite{ACT}.

However, as scene complexity and visual distractions increase, the performance of policies that excel in simple environments tends to deteriorate \cite{IVM}. Recent works have explored leveraging internet-scale pretrained knowledge from vision-language foundation models (VLFMs) to improve generalization in downstream robotic tasks \cite{robo-gpt}. While promising, these methods often lack fine-grained visual perception capabilities. Policies are typically guided only by high-level language instructions or affordance-based cues \cite{rekep, keypoint}, limiting their ability to extract task-specific semantics from visual observations. For example, a policy may understand the instruction \textit{``pick up the red block"}, but without fine-grained perception it cannot reliably determine the block’s precise location, orientation, or its spatial relation to surrounding obstacles. Meanwhile, research on vision-language models (VLMs) has shown that image resolution plays a critical role in sustaining robust visual understanding in complex environments. For example, LLaVA-Next \cite{llavanext} and Otter-HD \cite{otterhd} demonstrate that higher-resolution inputs can substantially improve the performance of previous works. However, the resulting increase in visual embedding dimensionality also incurs significant computational costs, particularly under resource-constrained robotic setups. These challenges raise two key questions: 1) How can visual information be fully exploited in robotic systems while maintaining feasible computational cost? 2) How can the pretrained knowledge of vision-language foundation models be transferred into equally fine-grained visual semantic representations to guide task-relevant perception?

To address these challenges, as illustrated in Fig.~\ref{Fig:Teaser}, we propose \textbf{ImitDiff}, a diffusion-based visuomotor imitation learning policy guided by fine-grained semantics within a dual-resolution workflow. \textbf{ImitDiff} transfers foundation-model priors into visual representations to improve performance under visual distractions and complex scenes. Specifically, we employ a VLM to identify task-relevant objects from user instructions and the visual scene. To operationalize this, we construct a real-time open-vocabulary detect-track-segment pipeline that robustly transforms high-level language instructions into pixel-level semantic masks throughout the manipulation process. These semantic masks guide a dual-resolution workflow that efficiently captures both global and local visual features, enhancing perception while maintaining compact visual embeddings. Finally, the semantically guided dual-resolution features are used to condition a consistency-driven \cite{consistency} diffusion transformer (DiT) action head, which bridges rich semantic perception with real-time action generation.

Within this framework, \textbf{ImitDiff} demonstrates robust performance across a range of fine-grained manipulation tasks, maintaining stability under increased scene complexity and visual distractions. We evaluate its performance and generalization across four simulated and four real-world manipulation tasks. With only 100 demonstrations, \textbf{ImitDiff} achieves a higher success rate than state-of-the-art imitation learning policies as well as VLMs-augmented robotic manipulation frameworks in distraction-free settings, with the performance gap widening as distractions intensify. We further design zero-shot generalization experiments under two challenging scenarios: 1) visual distraction generalization, where models are trained without distractions but tested with them, and 2) object generalization, where test objects are unseen during training. The results highlight \textbf{ImitDiff}’s superior robustness and adaptability, enabled by the integration of fine-grained priors from foundation models and the dual-resolution workflow. Additionally, we evaluate various denoising strategies in terms of success rate and inference efficiency. Our consistency-driven diffusion transformer action head achieves an order-of-magnitude speedup while maintaining competitive success. Ablation studies further validate the contribution of each component within the overall architecture.

Our key contributions are as follows: 1) we develop a real-time open-vocabulary detect-track-segment pipeline that robustly transforms high-level user instructions into pixel-level visual semantic masks, enabling fine-grained semantic guidance over task-relevant regions in the latent space; 2) we introduce an efficient dual-resolution visual enhancement workflow based on a dual-encoder architecture, which maximizes multi-scale visual information extraction while maintaining compact visual embeddings; 3) we implement a consistency-driven diffusion transformer action head that delivers an order-of-magnitude acceleration in inference while preserving competitive success rates. Together, these contributions establish a strong framework for distraction-robust and real-time visuomotor imitation learning policy.

\begin{figure*}[htbp]
    \centering
    \includegraphics[width=1.0\textwidth]{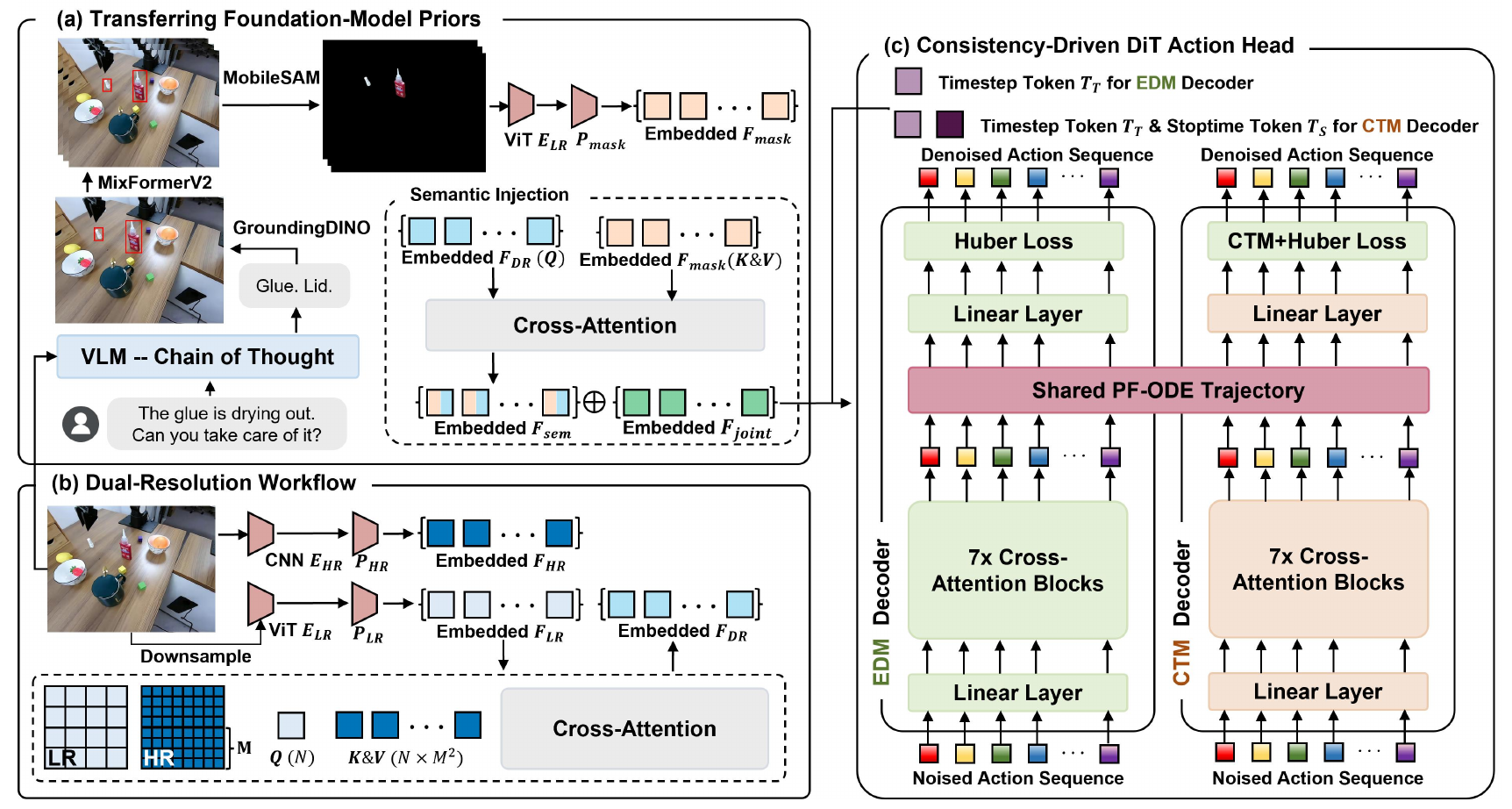} 
    \caption{Overview of \textbf{ImitDiff}. Our framework comprises three components: \textbf{a) Transferring Foundation-Model Priors.} Given a user instruction and the initial observation, a VLM identifies task-relevant objects through a chain-of-thought process. An open-vocabulary detect-track-segment pipeline then produces visual semantic masks, which are injected into a shared latent space via a semantic-injection encoder, thereby transferring foundation-model priors to the dual-resolution visual features. \textbf{b) Dual-Resolution Workflow.} For each camera view, both high and low resolution observations are obtained and encoded by a dual-encoder system. Low-resolution features query candidate regions within high-resolution features at the patch level via attention, maximizing multi-scale information extraction while retaining efficiency. \textbf{c) Consistency-Driven DiT Action Head.} An EDM action decoder is first trained as a teacher model conditioned on visual observations and proprioceptive inputs. A CTM student decoder is then distilled along the same PF-ODE trajectory, achieving substantially faster inference while preserving task success rates.}
    \label{Fig:Overview}
\vspace{-1.5\baselineskip}
\end{figure*}

\section{Related Work}
\label{Related}
\subsection{Visuomotor Imitation Learning}

Visuomotor imitation learning provides an effective framework for enabling robots to acquire human-like skills from expert video demonstrations \cite{UMI}. Recent advances in diffusion-based visuomotor policies have demonstrated strong potential in learning complex manipulation tasks by integrating visual observations with high-dimensional, multi-modal action distributions \cite{ACT}. However, most existing methods emphasize modeling action distributions with VAEs or diffusion models, while paying limited attention to visual perception itself \cite{ACT, diffusion-policy}. In contrast, \textbf{ImitDiff} introduces a dual-resolution workflow that maximizes the utility of visual information while maintaining compact visual embeddings.

\subsection{Vision-Language Foundation Models for Robotics}

Vision-language foundation models, such as VLMs \cite{gpt} and open-vocabulary object detectors \cite{groundingdino}, enable natural language to guide visual understanding through joint vision-language pretraining. These models demonstrate strong transferability in downstream robotics applications and often serve as semantic anchors for multi-modal representations, providing an intermediate grounding layer for planning and reasoning. Prior works such as VoxPoser \cite{voxposer} and Manipulate-Anything \cite{manipulate}, leverage the planning capabilities of VLMs to enhance the generalization of manipulation frameworks, typically adopting code-as-policy paradigms to decompose tasks into action primitives. Other methods, including ReKep \cite{rekep} and KALM \cite{keypoint}, exploit vision-language foundation models and pretrained vision priors (e.g., CLIP \cite{CLIP} and DINO \cite{dino}) to extract affordance cues from manipulation tasks. In contrast, \textbf{ImitDiff} advances this line of research by enabling a more fine-grained representation for semantic guidance: it transforms pretrained priors from vision-language models into pixel-level visual semantic masks that are modality-aligned with the input observations. These masks guide visual features to attend to task-relevant regions within a shared latent space.

\subsection{Acceleration Strategies for Diffusion Models in Robotics}

Diffusion models suffer from high inference latency due to their inherently iterative sampling process, posing a major challenge for real-time robotic control. Techniques such as Denoising Diffusion Implicit Models (DDIM) \cite{ddim} and Elucidated Diffusion Models (EDM) \cite{edm} reinterpret the sampling dynamics as an ordinary differential equation (ODE), enabling faster inference with fewer denoising steps at the cost of degraded sample quality. Other approaches, including Picard Iteration \cite{parallel}, exploit parallel computation to accelerate inference but remain impractical for resource-constrained robotic platforms. Shortcut-based methods, such as One-Step Diffusion \cite{one, wang2024one} and Inductive Moment \cite{zhou2025inductive}, approximate the diffusion trajectory with aggressive single-step or moment-matching strategies, achieving efficiency but often struggling with sample fidelity. In parallel, consistency-based distillation frameworks \cite{consistency} have shown that student models can take larger integration steps along the ODE trajectory defined by teacher models, striking a balance between inference speed and output quality. These works highlight the promise of consistency-based acceleration in real-time robot manipulation. Our contribution extends this line of research by being the first to couple a consistency-driven framework with a DiT architecture. Unlike prior U-Net–based policies \cite{prasad2024consistency}, the DiT action head excels at modeling long-horizon dependencies in action generation, enabling an order-of-magnitude speedup without sacrificing task performance.

\section{Method}
\label{Method}

Herein we discuss: 1) the motivation and formal problem definition; 2) how pretrained priors from foundation models are transformed into fine-grained visual semantic masks, and how these masks guide feature extraction in a dual-resolution workflow (Sec. \ref{Sec. III-B}); 3) the design and implementation of the dual-resolution workflow (Sec. \ref{Sec. III-C}) and 4) the motivation and architecture of the consistency-driven diffusion transformer action head (Sec. \ref{Sec. III-D}).

\subsection{Problem Formulation}
\label{Sec. III-A}

We aim to develop a generalizable and distraction-robust robotic system capable of interpreting high-level user instructions and executing precise manipulation actions, even under visually distracting conditions. For instance, in response to a command such as ``The glue is drying out, can you take care of it?", the robot autonomously performs the appropriate action ``covering the glue with the lid”. This illustrates how planning, perception, and control are tightly integrated within a unified visuomotor framework.

Thus, in our formulation, we define a visuomotor imitation learning policy $\pi_\varepsilon(a \mid p, o, l)$, where $\pi$ is a diffusion-based probabilistic model parameterized by $\varepsilon$. This policy maps proprioceptive input $p$, visual observation $o$, and user language instruction $l$ to continuous actions on a physical robot. To improve generalization and distraction robustness, the policy is composed of three key components: 1) transferring foundation-model priors (Sec. \ref{Sec. III-B}), which transforms high-level language instructions into fine-grained visual semantic masks and injects them into dual-resolution features to guide perception; 2) dual-resolution workflow (Sec. \ref{Sec. III-C}), which captures multi-scale visual features while maintaining compact visual representation; 3) consistency-driven DiT action head (Sec. \ref{Sec. III-D}), which significantly accelerates inference while preserving action accuracy.

\subsection{Transferring Foundation-Model Priors}
\label{Sec. III-B}

\textbf{Task-Relevant Object Reasoning.} As illustrated in Fig.~\ref{Fig:Overview}(a), we leverage the advanced VLM GPT-4o \cite{gpt} to infer task-relevant objects. Given a high-level user instruction (e.g., ``The glue is drying out, can you take care of it?") and an initial visual observation, GPT-4o performs structured reasoning using a carefully designed prompt template. It first generates a concise scene description, then infers the intended task (e.g., ``Cover the glue with the lid''), and finally filters out the relevant objects (glue and lid). This chain-of-thought inference not only allows the VLM to identify task-relevant objects accurately, but also introduces an explicit, high-level semantic filtering mechanism to suppress irrelevant distractions. This serves as the first layer of semantic grounding for robust visuomotor policy learning.

\textbf{Open-Vocabulary Detect-Track-Segment Pipeline.} Following task-relevant object reasoning, we adopt the state-of-the-art open-vocabulary detector GroundingDINO \cite{groundingdino} to localize target objects from language instructions. However, during robotic manipulation, these objects are frequently subject to occlusion and dynamic interference, making detection alone unreliable. To improve temporal robustness, we switch to the tracker MixFormerV2~\cite{mixformerv2} after acquiring the initial bounding boxes, leveraging its spatiotemporal continuity for consistent object localization. In robotic tasks, precise object masks with rich shape and geometric priors are crucial for modeling affordance-related constraints. To generate these masks in real time, we employ MobileSAM \cite{mobilesam}, which produces frame-wise RGB semantic masks from bounding-box prompts infused with semantic cues. This pipeline enables the high-level language instructions to be translated into pixel-level visual semantic masks in real time, forming the second layer of explicit, pixel-level filtering against visual distractions, based on an open-vocabulary foundation model pipeline.

\textbf{Semantic-Injection Encoder.} As shown in Fig.~\ref{Fig:Overview}(a), we process high-resolution observations using the vision-language foundation model pipeline to generate pixel-level visual semantic masks. This design leverages the superior spatial fidelity of high-resolution inputs to mitigate hallucinations from foundation models. To ensure alignment with the feature dimensions of the final dual-resolution representation (which matches that of the low-resolution query features in the dual-resolution workflow), we resize the semantic masks to the spatial resolution of the low-resolution input. The resized masks are denoted as ${O}_{\text{mask}} \in {R}^{H \times W \times 3}$. To guide visual features within a shared latent space, we use a ViT-based encoder $E_{\text{LR}}$ (weight-shared with the low-resolution branch) along with a semantic projector $P_{\text{mask}}$ to extract task-conditioned semantic features ${F}_{\text{mask}}$. These serve as keys and values in 4$\times$ cross-attention blocks, with the dual-resolution visual features $F_{\text{DR}}$ acting as queries, yielding the final task-aware features $F_{\text{sem}}$. This semantic fusion introduces a third layer of implicit, feature-level filtering against visual distractions. Together, the VLM reasoning module, open-vocabulary detect-track-segment pipeline, and semantic-injection encoder effectively transfer semantic, geometric, and spatiotemporal priors from foundation models into the end-to-end visuomotor policy.

\subsection{Dual-Resolution Workflow}
\label{Sec. III-C}

\textbf{Dual-Encoder System.} In our formulation, we aim to enhance the policy’s visual perception capability—specifically, the observation component $o$ in $\pi_\varepsilon(a \mid p, o, l)$. Inspired by dual-encoder architectures in computer vision \cite{doubly}, we design a dual-resolution workflow to adapt this paradigm for robotic manipulation. Our system consists of two parallel branches: a high-resolution stream and a low-resolution stream. The low-resolution input $O_{\text{LR}} \in {R}^{H \times W \times 3}$ is derived by downsampling the high-resolution input $O_{\text{HR}} \in {R}^{H' \times W' \times 3}$. In the low-resolution branch, we employ a CLIP-pretrained ViT encoder $E_{\text{LR}}$ to extract visual features $F_{\text{LR}} \in {R}^{N \times D}$, where $N$ denotes the number of visual patches and $D$ the embedding dimension. Attention across these patch tokens captures long-range dependencies, facilitating global visual context modeling. To ensure domain consistency with $E_{\text{LR}}$, we use a CLIP-pretrained ConvNeXt as the high-resolution encoder $E_{\text{HR}}$. A feature pyramid network (FPN) is constructed over the intermediate feature maps $F^i_{\text{HR\_map}}$ to capture multi-scale spatial information. Specifically, we apply $1 \times 1$ convolutions to unify the channel dimensions across stages, followed by progressive top-down upsampling and feature fusion. The bottom-most feature map retains spatial details while aggregating semantic cues from deeper layers. To mitigate upsampling artifacts, we apply a $3 \times 3$ convolution to the final fused output, producing the high-resolution feature map $F_{\text{HR}} \in {R}^{N' \times N' \times D}$. As illustrated in Fig.~\ref{Fig:Overview}(b), the spatial feature map $F_{\text{HR}}$ is reshaped into a sequence of $N \times M^2$ tokens, where each set of $M^2$ high-resolution patches is spatially aligned with one corresponding token in $F_{\text{LR}}$. This alignment forms the basis for the patch-level cross-attention mechanism used in subsequent fusion.

\textbf{Dual-Resolution Fusion.} Given the previously obtained low-resolution features $F_{\text{LR}}$ and high-resolution features $F_{\text{HR}}$, we perform patch-level cross-attention to effectively fuse global and local visual information. As shown in Fig.~\ref{Fig:Overview}(b), the low-resolution features $F_{\text{LR}}$ are used as queries $Q \in {R}^{N \times D}$, while the high-resolution features $F_{\text{HR}}$ serve as $K \in {R}^{N \times M^2 \times D}$ and $V \in {R}^{N \times M^2 \times D}$. Each low-resolution query attends only to its spatially aligned group of $M^2$ high-resolution patches, enabling localized attention within the corresponding region of $F_{\text{HR}}$. This targeted attention mechanism preserves computational efficiency by avoiding full-map interactions, while still allowing each token to retrieve fine-grained visual cues from high-resolution context. Through this fusion, we obtain a unified dual-resolution feature representation $F_{\text{DR}}$ that integrates the global semantic context from $F_{\text{LR}}$ with detailed spatial information from $F_{\text{HR}}$, maximizing perceptual expressiveness without increasing the embedding size.

\subsection{Consistency-Driven DiT Action Head}
\label{Sec. III-D}

The preceding modules yield a compact visual representation $F_{\text{sem}}$ (see semantic-injection encoder in Sec.~III-B), that is both semantically grounded and resolution-enhanced, capturing global task intent and fine-grained spatial details. While this embedding is well-suited for visuomotor imitation, the iterative sampling process of diffusion models poses a significant bottleneck for real-time robotic deployment.

To address this challenge, we propose a consistency-driven strategy that distills a lightweight student decoder from a fully denoising diffusion transformer. This approach bridges rich semantic perception with low-latency control, significantly reducing inference time while preserving the alignment between the action policy and task-relevant semantics.

\textbf{EDM Framework for Action Decoder.} As illustrated in Fig.~\ref{Fig:Overview}(c), we adopt the EDM framework to train a transformer-based teacher decoder $G_{\phi}$, which enables accurate and consistent action generation across diverse manipulation tasks. Specifically, we concatenate the visual semantic condition $F_{\text{sem}}$ with proprioceptive features $F_{\text{joint}}$ and a time-step embedding $T_t$ as the key and value inputs to the decoder, while the query consists of a noised action sequence $a_t$ sampled at diffusion step $t$. This design allows the decoder to model the denoising process of multi-modal action distributions conditioned on both the observation and diffusion timestep. The probability flow ordinary differential equation (PF-ODE) is estimated through attention-based interactions between the noisy actions and the observation-conditioned context:

\begin{equation}
dx_t/dt = -(x_t - G_\phi(x_t, t; o))/t
\end{equation}

To supervise training, we employ an optimized denoising score matching (DSM) loss, which samples along the PF-ODE trajectory $(a_t, t)$ and trains the decoder to recover the original action $a_0$.

\begin{equation}
L_{\text{DSM}}(\phi) = \mathrm{E}_{t, a_0, a_t \mid x_0} \left[ d(a_0, G_\phi(a_t, t; o)) \right]
\end{equation}

Here, $d$ denotes the optimized Huber loss:

\begin{equation}
d(x, y) = \sqrt{\|x - y\|_2^2 + c^2} - c
\end{equation}

\begin{figure}[t]
    \centering
    \includegraphics[width=\linewidth]{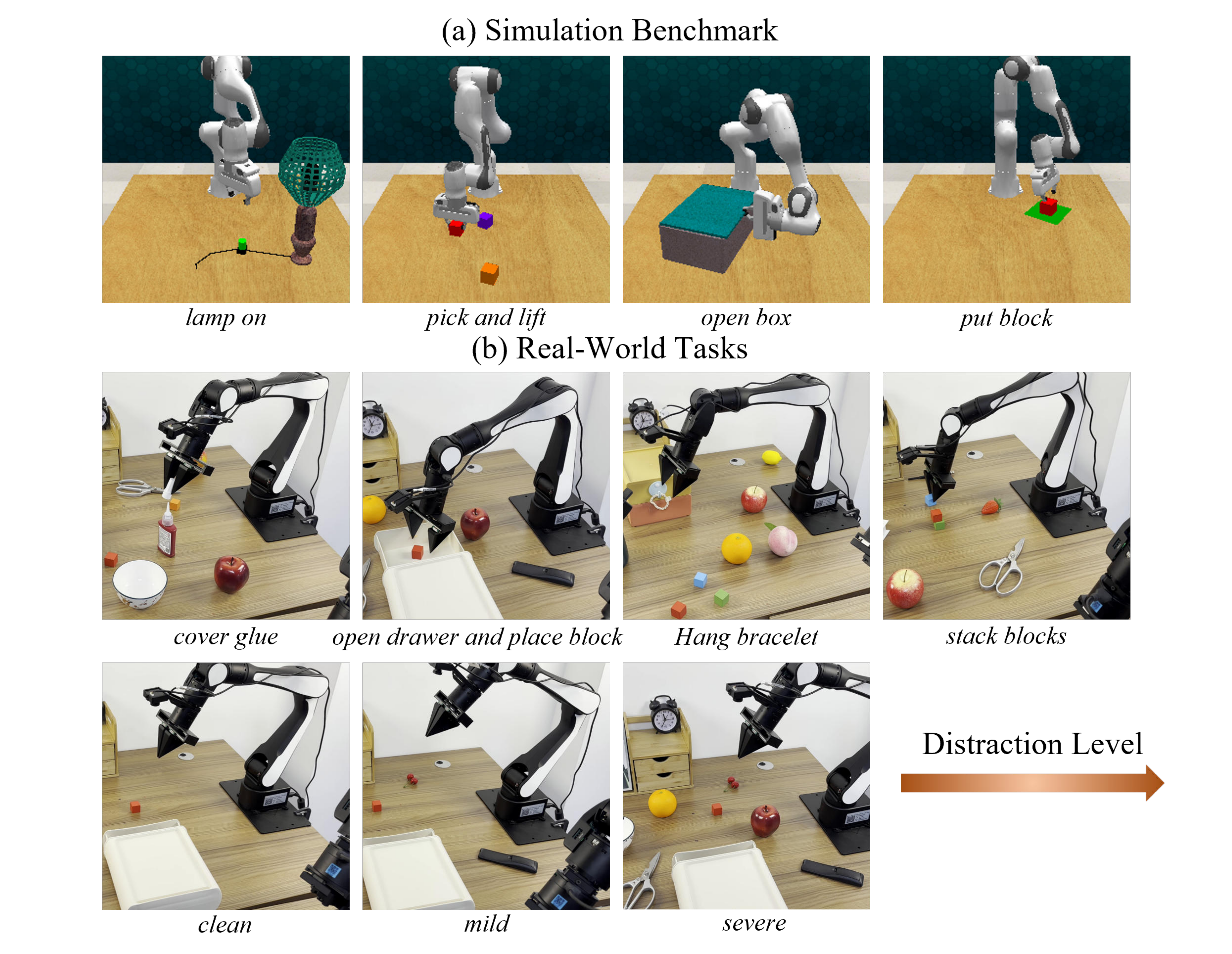} 
    \caption{Tasks in simulation and real-world experiments. Real-world trials include foreground distractors at increasing strengths.}
    \label{Fig:Tasks_V2}
\vspace{-1.5\baselineskip}
\end{figure}

\textbf{CTM Framework for Action Decoder.} While the EDM-based teacher decoder ensures accurate and consistent action generation, its iterative denoising process incurs significant inference latency. In time-critical robotic tasks, such latency impairs control responsiveness and consequently degrades task performance, especially in dynamic or visually distracting environments.

To overcome this limitation, we introduce a consistency trajectory model (CTM) that distills the EDM-based policy into a lightweight student decoder. CTM is trained to trace the same PF-ODE trajectory as the teacher, substantially reducing denoising steps while maintaining semantic alignment. Given the same semantically grounded embedding $F_{\text{sem}}$, the student decoder produces actions consistent with the teacher policy, enabling real-time execution.

We define the student model as $g_\phi(a_t, t, s; o)$. Similar to the EDM-based diffusion transformer decoder, it takes as input the observation condition $o$, the current time step $t$, and the target stopping time $s$. To model consistency along the same PF-ODE trajectory, we sample two noisy actions $(a_{t_1}, t_1)$ and $(a_{t_2}, t_2)$, and map them to a shared intermediate state at time $s$: $a_s^{t_1} = g_\phi(a_{t_1}, t_1, s; o)$ and $a_s^{t_2} = g_\phi(a_{t_2}, t_2, s; o)$. These are then further denoised to the initial step $t = 0$, yielding $g_\phi(a_s^{t_1}, s, 0; o)$ and $g_\phi(a_s^{t_2}, s, 0; o)$. The consistency loss is computed in the fully denoised action space to enforce trajectory-level alignment.

\begin{equation}
L_{\text{CTM}} = d\left(g_\phi(a_s^{t_1}, s, 0; o), g_\phi(a_s^{t_2}, s, 0; o)\right)
\end{equation}

The CTM framework is trained using a joint objective that combines $L_{\text{DSM}}$ and $L_{\text{CTM}}$.

\begin{equation}
L = \alpha L_{\text{CTM}} + \beta L_{\text{DSM}}
\end{equation}

\begin{figure}[t]
    \centering
    \includegraphics[width=1.0\linewidth]{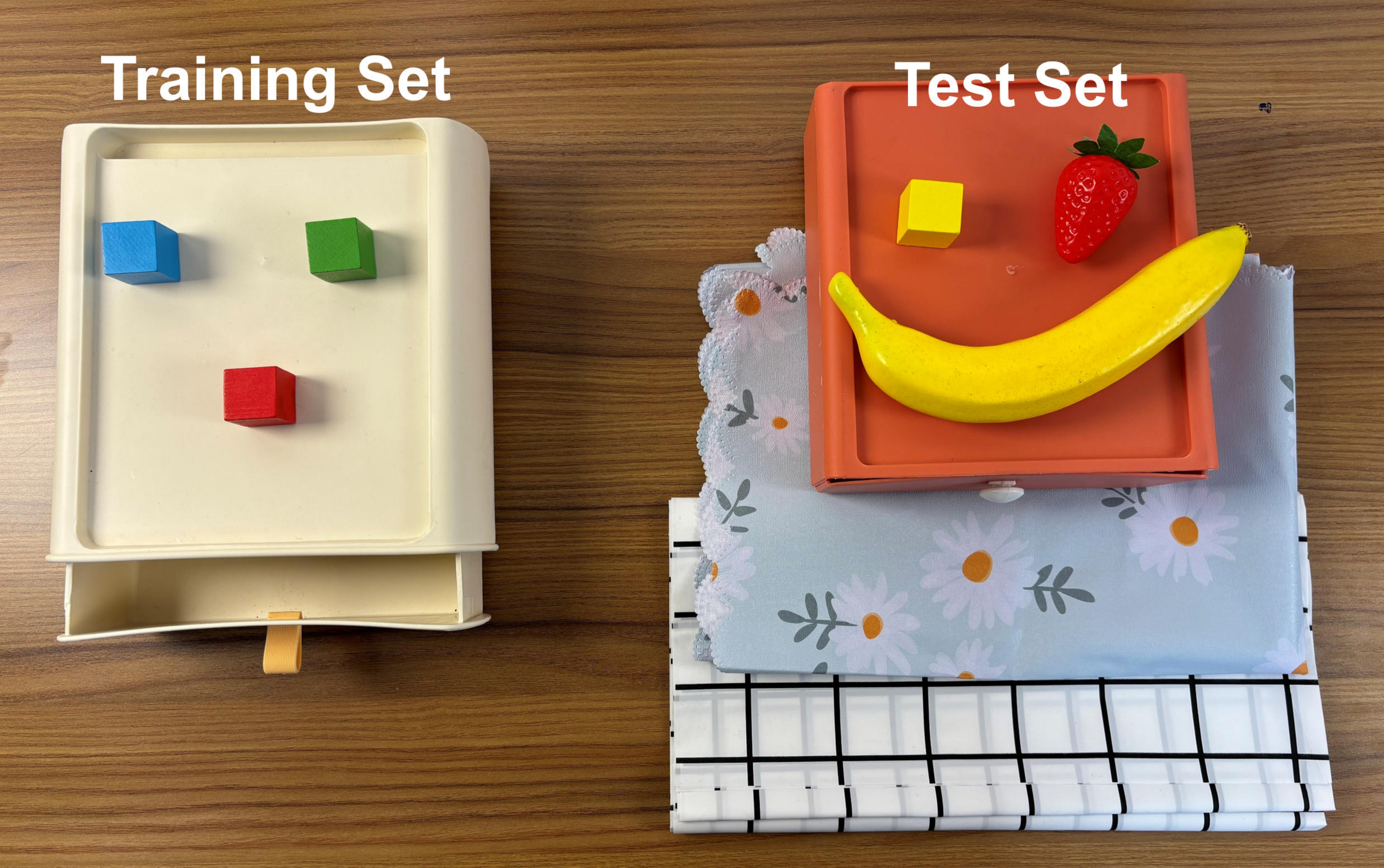} 
    \caption{Objects visualization for the training and test sets.}
    \label{Fig:Zero_Shot}
\vspace{-1.5\baselineskip}
\end{figure}

\section{Experiments}
\label{Experiments}

We assess the effectiveness of \textbf{ImitDiff} through a series of simulation and real-world robotic manipulation experiments. This section addresses the following key questions: 1) under what simulation setups is \textbf{ImitDiff} evaluated, and how does it perform (Sec. \ref{Sec. IV-A})? 2) under what real-world setups is \textbf{ImitDiff} evaluated, and how does it perform (Sec. \ref{Sec. IV-B})? 3) what roles do the individual components of \textbf{ImitDiff} play in the overall system performance (Sec. \ref{Sec. IV-C})?

\subsection{Simulation Experiments}
\label{Sec. IV-A}

\textbf{Environment and tasks.} We use RLBench, a robot learning benchmark based on CoppeliaSim for simulation evaluation \cite{james2020rlbench}. We sample 4 tasks from RLBench, covering a diverse range of action primitives and task horizons. All simulation experiments use a Franka Panda robot with a parallel gripper. Input observations are captured from two RGB cameras mounted at the front and the wrist. We present the tasks in the simulation experiments in Fig.~\ref{Fig:Tasks_V2}(a).

\textbf{Baselines.} We compare \textbf{ImitDiff} against two state-of-the-art robot manipulation frameworks integrated with vision-language foundation models: VoxPoser \cite{voxposer} and Manipulate-Anything (MA) \cite{manipulate}. VoxPoser builds a 3D voxel map of value functions for predicting waypoints. MA leverages vision-language foundation models to decompose tasks into sub-goals and generate 6-DoF action poses. We follow the experimental settings reported in the original MA work for VoxPoser and MA, and align the resolution of \textbf{ImitDiff} accordingly (128 for low-resolution and 256 for high-resolution). It should be noted that we set the denoising steps of CTM to 1 during all deployments.

\textbf{Results.} As shown in Tab.~\ref{Simulation_Results}, \textbf{ImitDiff} outperforms the baselines on 3 out of 4 tasks. This isolates the specific advantages of its perception module. Under a similar vision-language foundation model pipeline, ImitDiff benefits from finer-grained visual semantic masks and a dual-resolution workflow that shares the same modality as the semantic masks. The largest performance gain is observed in the \textit{open box} task, which requires multi-degree-of-freedom manipulation of the robot. This demonstrates that \textbf{ImitDiff} leverages the consistency-driven DiT action head to effectively learn action distributions that are difficult to capture with code-as-policy approaches.

\begin{table}[t]
\centering
\caption{Task-averaged success rate \% for simulation experiments.}
\label{Simulation_Results}
\begin{threeparttable}
\begin{tabular*}{\columnwidth}{@{\extracolsep{\fill}}lcccc@{}}
\toprule[0.8pt]
\textbf{Policy} & 
\textbf{Lamp On} & 
\textbf{Open Box} & 
\textbf{Pick and Lift} & 
\textbf{Put Block} \\
\midrule
\textbf{ImitDiff} & \textbf{89.33$\pm$1.25} & \textbf{62.00$\pm$5.10} & \textbf{96.67$\pm$0.47} & 92.33$\pm$2.62 \\
MA & 69.33$\pm$6.11 & 29.00$\pm$10.07 & 84.00$\pm$6.93 & \textbf{96.00$\pm$4.00} \\
VoxPoser & 57.30$\pm$12.22 & 0.00$\pm$0.00 & 96.00$\pm$0.00 & 70.70$\pm$2.31 \\
\bottomrule[0.8pt]
\end{tabular*}

\begin{tablenotes}[flushleft]
\footnotesize
\item Each task is evaluated over 3 seeds to obtain the task-averaged success rate and standard deviations on 100 rollouts with 100 demonstrations for training.
\end{tablenotes}
\end{threeparttable}
\vspace{-1.5\baselineskip}
\end{table}

\subsection{Real-World Experiments}
\label{Sec. IV-B}

\textbf{Environment and tasks.} Our experimental platform is built upon the AIRBOT Play robotic arms, with a teach pendant and a gripper for demonstration and inference. Input observations are captured from two RGB cameras mounted at the global and the wrist. Inference runs on a desktop equipped with a single NVIDIA 4060Ti GPU (16GB). We present the tasks in the real-world experiments in Fig.~\ref{Fig:Tasks_V2}(b). For each task, we collect 100 teleoperated demonstrations, comprising 50 under no distraction, 30 under mild distraction, and 20 under severe distraction conditions.

\textbf{Baselines.} We compare \textbf{ImitDiff} against three advanced visuomotor imitation learning baselines: 1) Action Chunking Transformer (ACT): a C-VAE-based policy that utilizes high-resolution visual inputs to learn fine-grained manipulation skills; 2) U-Net-Based Diffusion Policy (DP-C): a convolutional diffusion model that models the multi-modal distribution of robot actions from visual observations; 3) Transformer-Based Diffusion Policy (DP-T): a diffusion policy variant that replaces the convolutional decoder with a transformer-based architecture to better model long-range temporal dependencies in trajectory prediction. We follow the real-world experimental setups reported in the original works of ACT \cite{ACT} and DP \cite{UMI}, and align the resolution of ImitDiff accordingly (224 for low-resolution and 448 for high-resolution).

\textbf{Main Results.} We comprehensively evaluate \textbf{ImitDiff} across four real-world robotic manipulation tasks, spanning a range of skills from basic
\textit{pick and place} to more challenging operations such as insertion and articulated-object handling. As shown in Tab.~\ref{Real_World_Experiments}, with only 100 demonstrations, \textbf{ImitDiff} achieves a 20\% higher success rate than the strongest baseline under distraction-free conditions, with the performance gap further widening as visual distractions intensify. Both baseline methods rely on VAE or diffusion-based approaches to model action distributions under the condition of observation features. These approaches share a similar design with \textbf{ImitDiff}'s consistency-driven DiT action head, which allows us to more clearly isolate and validate the specific advantages of \textbf{ImitDiff}’s perception module. The results indicate that the observed performance gains primarily stem from the perception stack, which integrates a dual-resolution workflow with pretrained priors from foundation models.

\textbf{Zero-Shot Generalization under Visual Distractions.} To further assess each \textbf{ImitDiff}’s ability to generalize to unseen visual distractions, we design a zero-shot test scenario that simulates real-world deployment conditions with varying environmental distractions. Specifically, we re-collected 100 demonstrations of the \textit{open drawer and place block} task (chosen to evaluate under foreground distractions, since this task is more sensitive to changes in the foreground) and 100 demonstrations of \textit{stack blocks} (chosen to evaluate under background distractions, as this task is more sensitive to variations in the background), all under distraction-free conditions and used them to train each policy. We then introduced varying levels of foreground and background visual distractions as illustrated in Fig.~\ref{Fig:Zero_Shot}. As shown in Tab.~\ref{tab:zero-shot_1}, \textbf{ImitDiff} consistently outperforms ACT and DP under different degrees and types of unseen visual distractions. This advantage stems from the open-vocabulary detect–track–segment pipeline integrated into \textbf{ImitDiff}, which leverages broad prior knowledge to identify target objects even in unseen environments. By injecting these visual priors into policy reasoning in the form of semantic masks, \textbf{ImitDiff} achieves enhanced robustness against novel visual distractions.

\textbf{Zero-Shot Generalization to Unseen Objects.} To further evaluate \textbf{ImitDiff}'s adaptability to variations in task objects, we design a zero-shot generalization experiment where all evaluation objects are excluded during training. This setting mimics real-world deployment scenarios where robots must manipulate objects with novel appearances or categories. Specifically, in the \textit{open drawer and place block} task, we train using one set of standard objects and evaluate on a distinct set of unseen target objects at test time, which differ in shape, size, and color as illustrated in Fig.~\ref{Fig:Zero_Shot}. This setup ensures that the assessment of generalization focuses on variations in the objects themselves. As shown in Tab.~\ref{tab:zero-shot_2}, despite the complete absence of these objects during training, \textbf{ImitDiff} consistently achieves significantly higher success rates than baseline policies across all test tasks. This advantage arises from two key factors: 1) the dual-resolution workflow extracts visual features at different scales, reducing \textbf{ImitDiff}’s reliance on specific textures and appearances; and 2) vision-language foundation models possess cross-task semantic reasoning abilities, enabling semantic recognition of novel objects and guiding \textbf{ImitDiff}’s perception layer to focus attention on them via semantic masks, thereby enhancing its generalization to unseen task objects.

\begin{table*}[t]
\centering
\begin{threeparttable}
\caption{Task-averaged success rate \% for real robot experiments.}
\label{Real_World_Experiments}
\begin{tabular*}{\textwidth}{@{\extracolsep{\fill}}lccccccccccccccc@{}}
\toprule[0.8pt]
\multirow{2}{*}{\textbf{Policy}} & 
\multicolumn{3}{c}{\makecell[c]{\textbf{Cover Glue}}} & 
\multicolumn{3}{c}{\makecell[c]{\textbf{Open Drawer}\\\textbf{and Place Block}}} &
\multicolumn{3}{c}{\makecell[c]{\textbf{Hang Bracelet}}} & 
\multicolumn{3}{c}{\makecell[c]{\textbf{Stack Blocks}}} & 
\multicolumn{3}{c}{\makecell[c]{\textbf{Average}}} \\
\cmidrule(lr){2-4}
\cmidrule(lr){5-7}
\cmidrule(lr){8-10}
\cmidrule(lr){11-13}
\cmidrule(lr){14-16}
& Clean & Mild & Severe & Clean & Mild & Severe & Clean & Mild & Severe & Clean & Mild & Severe & Clean & Mild & Severe \\
\midrule
\textbf{ImitDiff} & \textbf{23/25} & \textbf{23/25} & \textbf{21/25} 
& \textbf{22/25} & \textbf{22/25} & \textbf{20/25} 
& \textbf{21/25} & \textbf{21/25} & \textbf{19/25} 
& \textbf{21/25} & \textbf{20/25} & \textbf{18/25} 
& \textbf{0.87} & \textbf{0.86} & \textbf{0.78} \\
ACT & 18/25 & 14/25 & 4/25 
& 16/25 & 13/25 & 4/25 
& 19/25 & 14/25 & 6/25 
& 14/25 & 9/25 & 0/25 
& 0.67 & 0.50 & 0.14 \\
DP-T & 17/25 & 8/25 & 0/25 
& 14/25 & 6/25 & 0/25 
& 17/25 & 10/25 & 0/25 
& 11/25 & 6/25 & 0/25 
& 0.59 & 0.30 & 0 \\
DP-C & 17/25 & 8/25 & 0/25 
& 13/25 & 6/25 & 0/25 
& 17/25 & 11/25 & 0/25 
& 12/25 & 5/25 & 0/25 
& 0.59 & 0.30 & 0 \\
\bottomrule[0.8pt]
\end{tabular*}

\begin{tablenotes}[flushleft]
\footnotesize
\item Each task is evaluated on 25 rollouts.
\end{tablenotes}
\end{threeparttable}
\vspace{-1.5\baselineskip}
\end{table*}

\begin{table}[t]
\centering
\begin{threeparttable}
\caption{Zero-shot experiments for visual distraction.}
\label{tab:zero-shot_1}

\begin{tabular*}{\linewidth}{@{\extracolsep{\fill}}lcccccc@{}}
\toprule[0.8pt]
\multirow{2}{*}{\textbf{Policy}} & 
\multicolumn{3}{c}{\textbf{(a) Foreground}} & 
\multicolumn{3}{c}{\textbf{(b) Background}} \\
\cmidrule(lr){2-4}\cmidrule(lr){5-7}
& Clean & Mild & Severe & Origin & Texture 1 & Texture 2 \\
\midrule
\textbf{ImitDiff} & \textbf{22/25} & \textbf{20/25} & \textbf{15/25} & \textbf{21/25} & \textbf{15/25} & \textbf{15/25} \\
ACT               & 16/25 & 10/25 & 0/25  & 14/25 & 6/25  & 5/25  \\
DP-T              & 14/25 & 4/25  & 0/25  & 11/25 & 4/25  & 5/25  \\
DP-C              & 13/25 & 4/25  & 0/25  & 12/25 & 5/25  & 5/25  \\
\bottomrule[0.8pt]
\end{tabular*}

\begin{tablenotes}[flushleft]
\footnotesize
\item Foreground distraction varies the number of distractor objects; background distraction changes the surrounding background. Each task is evaluated on 25 rollouts.
\end{tablenotes}

\end{threeparttable}
\vspace{-1.5\baselineskip}
\end{table}

\subsection{Ablation Studies}
\label{Sec. IV-C}

To assess the individual contributions of each component in \textbf{ImitDiff}, we perform a comprehensive ablation study on the \textit{cover glue} task, due to its stringent requirements for manipulation precision. The results are summarized in Tab.~\ref{Ablation}. We provide a detailed analysis of how each module affects overall performance as follows:

\textbf{Pretrained Visual Encoders.} We replace the original CLIP-pretrained encoders in the dual-resolution workflow with a DINO-pretrained ViT for the low-resolution branch and an ImageNet-pretrained ConvNeXt for the high-resolution branch. As shown in Rows 1 and 2 of Tab.~\ref{Ablation}, both configurations achieve similar performance, demonstrating that the superior performance of \textbf{ImitDiff} stems from its architectural design itself, rather than from reliance on specific pretrained visual encoder weights.

\textbf{Visual Semantic Masks.} Removing the visual semantic masks leads to a significant performance drop under visual distractions, as shown in Row 3 of Tab.~\ref{Ablation}, with success rates falling from 88\% to 48\% in severe cases. This highlights the importance of the semantic masks, which transfer foundation-model priors to guide attention toward task-relevant regions and suppress distractions in the latent space.

\textbf{Feature Pyramid Network.} As indicated in Row 4 of Tab.~\ref{Ablation}, removing the feature pyramid network degrades performance across all distraction levels. This confirms its essential role in extracting multi-scale spatial features from high-resolution observations, which are especially critical for robust and precise visual understanding.

\begin{table}[!t] 
\centering 
\begin{threeparttable} 
\caption{Zero-shot experiments for unseen objects.} \label{tab:zero-shot_2} 
\begin{tabular*}{\columnwidth}{@{\extracolsep{\fill}}lccc@{}} \toprule[0.8pt] \multirow{2}{*}{\textbf{Policy}} & \multicolumn{3}{c}{\textbf{Unseen Objects}} \\ \cmidrule(lr){2-4} & Shape Shift & Size Shift & Color Shift \\ \midrule \textbf{ImitDiff} & \textbf{19/25} & \textbf{17/25} & \textbf{19/25} \\ ACT & 11/25 & 11/25 & 13/25 \\ DP-T & 8/25 & 6/25 & 8/25 \\ DP-C & 8/25 & 6/25 & 9/25 \\ \bottomrule[0.8pt] 
\end{tabular*} 
\begin{tablenotes}[flushleft] 
\footnotesize 
\item Each task is evaluated on 25 rollouts. Unseen factors: shape, size, color. 
\end{tablenotes} 
\end{threeparttable} 
\vspace{-1.5\baselineskip}
\end{table}

\begin{table*}[!t]
\centering
\caption{Ablation Study on Components of ImitDiff}
\label{Ablation}
\begin{tabular*}{\textwidth}{@{\extracolsep{\fill}}ccccccccc@{}}
\toprule[0.8pt]
\textbf{Low Res} & \textbf{High Res} & \textbf{Multi-Scale} & \textbf{Semantic Mask} & \textbf{Pretrained-Domain} & \textbf{Clean-Distraction} & \textbf{Mild-Distraction} & \textbf{Severe-Distraction} \\
\midrule
\textcolor{green!60!black}{\ding{51}} & \textcolor{green!60!black}{\ding{51}} & \textcolor{green!60!black}{\ding{51}} & \textcolor{green!60!black}{\ding{51}} & DINO-Pretrained & \textbf{23/25} & 22/25 & \textbf{22/25} \\
\textcolor{green!60!black}{\ding{51}} & \textcolor{green!60!black}{\ding{51}} & \textcolor{green!60!black}{\ding{51}} & \textcolor{green!60!black}{\ding{51}} & CLIP-Pretrained & \textbf{23/25} & \textbf{23/25} & 21/25 \\
\textcolor{green!60!black}{\ding{51}} & \textcolor{green!60!black}{\ding{51}} & \textcolor{green!60!black}{\ding{51}} & \textcolor{red!70!black}{\ding{55}} & CLIP-Pretrained & 22/25 & 19/25 & 12/25 \\
\textcolor{green!60!black}{\ding{51}} & \textcolor{green!60!black}{\ding{51}} & \textcolor{red!70!black}{\ding{55}} & \textcolor{green!60!black}{\ding{51}} & CLIP-Pretrained & 19/25 & 16/25 & 15/25 \\
\textcolor{red!70!black}{\ding{55}} & \textcolor{green!60!black}{\ding{51}} & \textcolor{green!60!black}{\ding{51}} & \textcolor{green!60!black}{\ding{51}} & CLIP-Pretrained & 18/25 & 18/25 & 16/25 \\
\textcolor{green!60!black}{\ding{51}} & \textcolor{red!70!black}{\ding{55}} & \textcolor{red!70!black}{\ding{55}} & \textcolor{green!60!black}{\ding{51}} & CLIP-Pretrained & 16/25 & 15/25 & 12/25 \\
\textcolor{red!70!black}{\ding{55}} & \textcolor{red!70!black}{\ding{55}} & \textcolor{red!70!black}{\ding{55}} & \textcolor{green!60!black}{\ding{51}} & CLIP-Pretrained & 4/25 & 2/25 & 0/25 \\ 
\bottomrule[0.8pt]
\end{tabular*}
\vspace{-1.5\baselineskip}
\end{table*}

\begin{table}[htbp]
\centering
\begin{threeparttable}
\caption{Benchmarking of Perception Stack and Action Head.}
\label{denoise}

\begin{tabularx}{\columnwidth}{@{}l *{4}{>{\centering\arraybackslash}X}@{}}
\toprule[0.8pt]
\makecell[c]{\textbf{Denoising Strategies}} & \textbf{DDPM} & \textbf{DDIM} & \textbf{EDM} & \textbf{CTM} \\
\midrule
Success Rate          & 83.33$\pm$9.43 & 80.00$\pm$8.16 & \textbf{93.33$\pm$4.71} & 90.00$\pm$8.16 \\
Encoder Time (ms)     & 27              & 27              & 27              & 27             \\
Action Head Time (ms) & 197             & 21              & 201             & \textbf{8.57}           \\
\bottomrule[0.8pt]
\end{tabularx}

\begin{tablenotes}[flushleft]\footnotesize
\item The task is evaluated over 3 seeds to obtain the task-averaged success rate and standard deviations on 10 rollouts. Inference time of encoder and action head is tested on GPU 4060Ti.
\end{tablenotes}

\end{threeparttable}
\vspace{-1.5\baselineskip}
\end{table}

\textbf{High-Resolution Workflow.} In Row 5, we ablate the entire high-resolution stream, including the feature pyramid. This results in a considerable decline in performance, indicating that the fine-grained spatial information and geometric details provided by the high-resolution branch are fundamental to accomplishing all manipulation tasks.

\textbf{Low-Resolution Workflow.} In Row 6, we remove the low-resolution stream and instead fuse the high-resolution features directly with resized semantic inputs. This ablation clearly reveals the low-resolution stream’s importance in modeling global visual context, which complements the spatial specificity of the high-resolution branch.

\textbf{Only Use Semantic Mask.} The results in row 7 show that relying solely on semantic masks makes it nearly impossible to complete the tasks. This is because semantic masks only indicate the pixel regions of the target objects, without providing information about the overall workspace layout or the relative spatial relationships between regions. Moreover, the absence of geometric details from the high-resolution branch further limits the model’s ability to extract task-relevant information, leading to a substantial performance drop.

\textbf{Denoising Strategies.} We benchmark different denoising strategies on the \textit{open drawer and place block} task under distraction-free conditions. Tab.~\ref{denoise} quantitatively reports the success rates and inference times under different denoising strategies. We find that CTM achieves a comparable success rate to EDM while offering significantly faster inference, demonstrating that the distillation process accelerates inference without compromising policy quality. In addition, we report the inference cost of the perception stack, and the quantitative results show that the perception module of \textbf{ImitDiff} incurs a reasonable computational overhead.

Our ablation studies clearly demonstrate that each component of the proposed \textbf{ImitDiff} framework contributes critically to the successful execution of complex robotic manipulation tasks. The highest performance is achieved when all components are integrated, underscoring the importance of their joint design and cohesive integration within the overall architecture.

\section{Conclusion}
\label{Conclusion}

In this work, we present \textbf{ImitDiff}, a distraction-robust visuomotor imitation learning framework that transfers foundation model priors into robotic policy learning. We leverage a vision-language foundation model pipeline to convert high-level user instructions into fine-grained visual semantic masks, which guide a dual-resolution visual workflow that efficiently integrates global context and local detail while preserving compact embeddings. These task-conditioned visual features are further used to condition a consistency-driven diffusion transformer (DiT) action head, enabling real-time, semantically aligned control. Extensive experiments and ablation studies confirm the effectiveness and generalization capability of our method. Future work will explore integrating more advanced foundation models to further improve policy generalization across diverse and unstructured environments.

\bibliographystyle{IEEEtran}
\bibliography{Reference}

\end{document}